\documentclass[utf8]{frontiersSCNS} % for Science, Engineering and 
\usepackage{url,lineno,microtype,subcaption}
\usepackage[hidelinks]{hyperref}
\usepackage[onehalfspacing]{setspace}
\usepackage{cite}
\usepackage{multirow}
\usepackage{array}
\usepackage{amsmath,graphicx,float}
\usepackage{afterpage}
\usepackage[ruled,vlined]{algorithm2e}
\usepackage{booktabs}
\usepackage{caption}
\usepackage{balance}
\usepackage{diagbox}
\graphicspath{{figure/}}
\DeclareMathOperator*{\argmax}{arg\,max}

\def\keyFont{\fontsize{8}{11}\helveticabold }
\def\firstAuthorLast{Wu {et~al.}} %use et al only if is more than 1 author
\def\Authors{Jibin Wu\,$^{1,*}$,  Emre Y{\i}lmaz\,$^{1}$, Malu Zhang\,$^{1}$, Haizhou~Li\,$^{1}$ and Kay~Chen~Tan\,$^{2}$}
\def\Address{$^{1}$Department of Electrical and Computer Engineering, National University of Singapore, Singapore\\
	$^{2}$Department of Computer Science, City University of Hong Kong, Hong Kong, China}
\def\corrAuthor{Jibin~Wu}
\def\corrEmail{jibin.wu@u.nus.edu}

\SetKwInput{KwInput}{Input}                % Set the Input
\SetKwInput{KwOutput}{Output}              % set the Output
\SetKwInput{KwFw}{Forward Pass}       % Set the Forward Pass
\SetKwInput{KwBw}{Backward Pass}       % Set the Backward Pass
\SetKwInput{KwLoss}{Loss}       % Set the Backward Pass
\SetKwInput{KwNote}{Note}       % Set the Backward Pass

\begin{document}
\onecolumn
\firstpage{1}

\title[Deep SNNs for Large Vocabulary ASR]{Deep Spiking Neural Networks for Large Vocabulary Automatic Speech Recognition} 

\author[\firstAuthorLast ]{\Authors} %This field will be automatically populated
\address{} %This field will be automatically populated
\correspondance{} %This field will be automatically populated

\extraAuth{}% If there are more than 1 corresponding author, comment this line and uncomment the next one.
%\extraAuth{corresponding Author2 \\ Laboratory X2, Institute X2, Department X2, Organization X2, Street X2, City X2 , State XX2 (only USA, Canada and Australia), Zip Code2, X2 Country X2, email2@uni2.edu}

\maketitle

\begin{abstract}
\section{}
Artificial neural networks (ANN) have become the mainstream acoustic modeling technique for large vocabulary automatic speech recognition (ASR). A conventional ANN features a multi-layer architecture that requires massive amounts of computation. The brain-inspired spiking neural networks (SNN) closely mimic the biological neural networks and can operate on low-power neuromorphic hardware with spike-based computation. Motivated by their unprecedented energy-efficiency and rapid information processing capability, we explore the use of SNNs for speech recognition. In this work, we use SNNs for acoustic modeling and evaluate their performance on several large vocabulary recognition scenarios. The experimental results demonstrate competitive ASR accuracies to their ANN counterparts, while require significantly reduced computational cost and inference time. Integrating the algorithmic power of deep SNNs with energy-efficient neuromorphic hardware, therefore, offer an attractive solution for ASR applications running locally on mobile and embedded devices.

\tiny
\keyFont{\section{Keywords:} Deep Spiking Neural Networks, Automatic Speech Recognition, Tandem Learning, Neuromorphic Computing, Acoustic Modeling}
\end{abstract}

\section{Introduction}
\label{sec:intro}
Automatic speech recognition (ASR) has enabled the voice interface of mobile devices and smart home appliances in our everyday life. The rapid progress in the integration of voice interfaces has been viable on account of the remarkable performance of the ASR systems using artificial neural networks (ANN) for acoustic modeling~\citep{lippmann1989,lang1990,hinton2012deep,YuDeng15}. Various ANN architectures, either feedforward or recurrent, have been investigated for modeling the acoustic information preserved in speech signals~\citep{dahl2012,graves2013,abdelhamid2014}.

The performance gains come with immense computational requirements often due to the time-synchronous processing of input audio signals. Several techniques have been proposed to reduce the computational load and memory storage of ANNs by reducing the number of parameters that have to be used for inference \citep{sainath2013,xue2013,he2014,povey2018}. Another common solution, for reducing the amount of processed speech, uses a wake word or phrase to initialize the embedded ASR engine and starts listening to input speech~\citep{zehetner2014,Sainath2015ConvolutionalNN,wu2018wakeword}. Moreover, most devices with voice control rely on cloud-based ASR engines rather than local on-device solutions. The necessity of online processing of speech via cloud computing comes with various concerns, such as data security and processing speed, etc. There have been multiple efforts to develop on-device ASR solutions in which the speech signal is processed locally using the computational resources of mobile devices~\citep{Lei2013AccurateAC,mcgraw2016}. 

Alternatively, event-driven models such as spiking neural networks (SNNs) inspired by the human brain have attracted ever-growing attention in recent years. The human brain is remarkably efficient and capable of performing complex perceptual and cognitive tasks. Notably, the adult's brain only consumes about 20 watts to solve complex tasks that are equivalent to the power consumption of a dim light bulb \citep{laughlin2003communication}. While brain-inspired ANNs have demonstrated great capabilities in many perceptual \citep{he2016deep, xiong2017toward} and cognitive tasks \citep{silver2017mastering}, these models are computationally intensive and memory inefficient to operate as compared to the biological brains. Unlike ANNs, asynchronous and event-driven information processing of SNNs resembles the computing paradigm that observed in the human brains, whereby the energy consumption matches the activity levels of  sensory stimuli. Given temporally sparse information transmitted in the surrounding environment, the event-driven computation, therefore, exhibits great computational efficiency than the synchronous computation used in ANNs.

Neuromorphic computing (NC), as a non-von Neumann computing paradigm, mimics the event-driven computation of the biological neural systems with SNN in silicon. The emerging neuromorphic computing architectures \citep{davies2018loihi, merolla2014million, furber2012overview} leverage on the massively parallel, low-power computing units to support spike-based information processing. Notably, the design of co-located memory and computing units effectively circumvents the von Neumann bottleneck of low-bandwidth between memory and the processing units \citep{monroe2014neuromorphic}. Therefore, integrating the algorithmic power of deep SNNs with the compelling energy efficiency of NC hardware represents an intriguing solution for pervasive machine learning tasks and always-on applications. Furthermore, growing research efforts are devoted to developing novel non-volatile memory devices for ultra-low-power implementation of biological synapses and neurons \citep{tang2019bridging}.

Some preliminary work on SNN-based phone classification or small-vocabulary speech recognition systems have been explored in \citep{liaw1998,nager2002,loiselle2005,holmberg2005,kroger2009,ref72,ref73,wu2018ijcnn,wu2018spiking,zhang2015digital,zhang2019mpd,bellec2018long}. However, these SNN-based ASR systems are far from the scale and complexity of  modern commercialized ANN-based ASR systems. It is mainly due to lacking effective training algorithms for deep SNNs and efficient software toolbox for SNN-based ASR systems. 

Due to the discrete and non-differentiable nature of spike generation, the powerful error back-propagation algorithm is not directly applicable to the training of deep SNNs. Recently, considerable research efforts are devoted to addressing this problem and the resulting learning rules can be broadly categorized into the SNN-to-ANN conversion \citep{cao2015spiking,diehl2015fast}, back-propagation through time with surrogate gradient  \citep{neftci2019surrogate,wu2018direct} and tandem learning \citep{hybridlearning}. Despite several successful attempts on the large-scale image classification tasks with deep SNNs \citep{ethImageNet,hu2018spiking,sengupta2019going,hybridlearning}, their applications to the large-vocabulary continuous ASR (LVCSR) tasks remain unexplored. In this work, we explore an SNN-based acoustic model for LVCSR using a recently proposed tandem learning rule \citep{hybridlearning} that supports an efficient and rapid inference. The experiments on several large vocabulary ASR benchmarks have shown competitive classification accuracies over the baseline ANN models. To the best of our knowledge, this is the first work that successfully applied SNNs for LVCSR tasks. Our preliminary study of an SNN-based acoustic model has also revealed compelling prospects of rapid inference and unprecedented energy efficiency of a neuromorphic approach.

% EMRE: This part is repeated at Section 5.2
% The emergence and active development of open-source software toolboxes, for instance, Kaldi \citep{povey2011kaldi} and ESPnet \citep{watanabe2018espnet}, play a significant role in the rapid progress of ANN-based ASR systems. In this work, we demonstrate that SNN-based acoustic models can be effectively developed in PyTorch and easily integrated into the PyTorch-Kaldi Speech Recognition Toolkit \citep{pytorch-kaldi} for rapid development of SNN-based ASR systems. 

The rest of the paper is organized as follows: In Section \ref{sec:relatedworks}, we first give an overview of spiking neural networks, large vocabulary ASR systems, and existing SNN-based ASR systems. In Section \ref{sec:method}, we introduce the spiking neuron model and the neural coding scheme that converts acoustic features into spike-based representation. We further present a recently introduced tandem learning framework for SNN training and how it is used to train deep SNN-based acoustic models. In Section~\ref{sec:result}, we present experimental results on the learning capability and energy efficiency of SNN-based acoustic models across three different types of recognition tasks including phone recognition, low-resourced and standard large-vocabulary ASR, and compare those to the ANN-based implementations. Finally, a discussion on the experimental findings is given in Section~\ref{sec:dis}.

\section{Fundamentals and Related Work}
\label{sec:relatedworks}
\subsection{Spiking Neural Networks}
The third generation spiking neural networks are originally studied as models to describe the information processing in the biological neural networks, wherein the information is communicated and exchanged via stereotypical action potentials or spikes \citep{gerstner2002spiking}. Neuroscience studies reveal that the temporal structure and frequency of these spike trains are both important information carriers in the biological neural networks. As will be introduced in Section \ref{subsec:neuronmodel}, the spiking neuron operates asynchronously and integrates the synaptic current from its incoming spike trains. An output spike is generated from the spiking neuron whenever its membrane potential crosses the firing threshold, and this output spike will be propagated to the connected neurons via the axon.

\begin{figure}[htb]
	\centering
	\centerline
	{\includegraphics[width = 9cm]{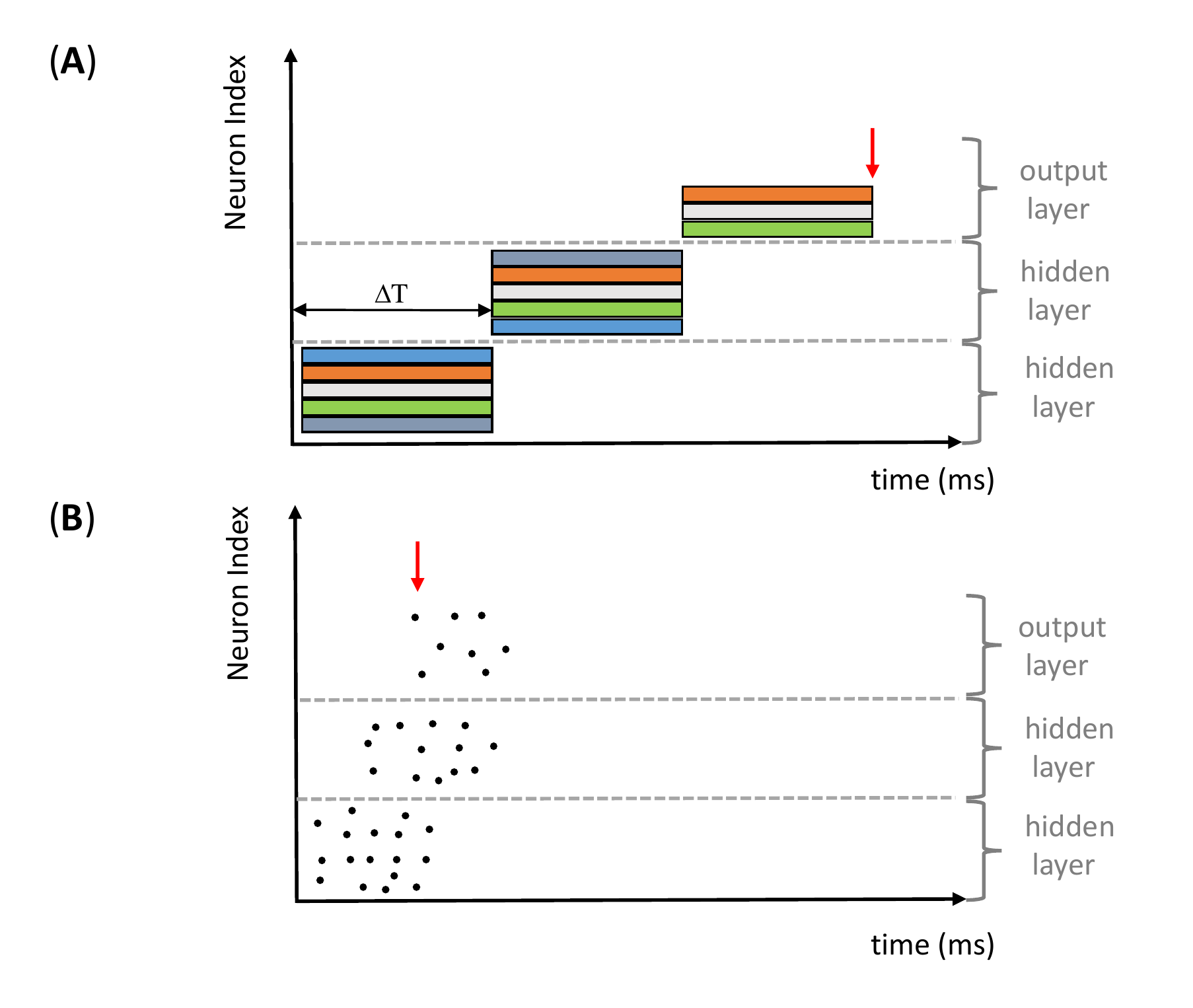}}
	\caption{Comparison of the synchronous and asynchronous computational paradigms adopted by (A) ANNs and (B) SNNs, respectively (revised from \citep{pfeiffer2018deep}).}
	\label{fig2}
\end{figure}

Motivated by the same connectionism principle, SNNs share the same network architectures, either feedforward or recurrent, with the conventional ANNs that use analog neurons. As shown in Figure~\ref{fig2}, the early classification decision can be made from the SNN since the generation of the first output spike. However, the quality of the classification decision is typically improved over time with more evidence accumulated. It differs significantly from the synchronous information processing of the conventional ANNs, where the output layer needs to wait until all preceding layers are fully updated. Therefore, despite information is transmitted and processed at a speed that is several orders of magnitude slower in neural substrates than signal processing in modern transistors, biological neural systems can perform complex tasks rapidly. For more overviews about SNNs and their applications, we refer readers to  \citep{pfeiffer2018deep, tavanaei2018deep}.

\subsection{Large Vocabulary Automatic Speech Recognition}
\label{ssec:lvasr}
\begin{figure}[htb]	
	\centering
	\centerline
	{\includegraphics[width = 12cm]{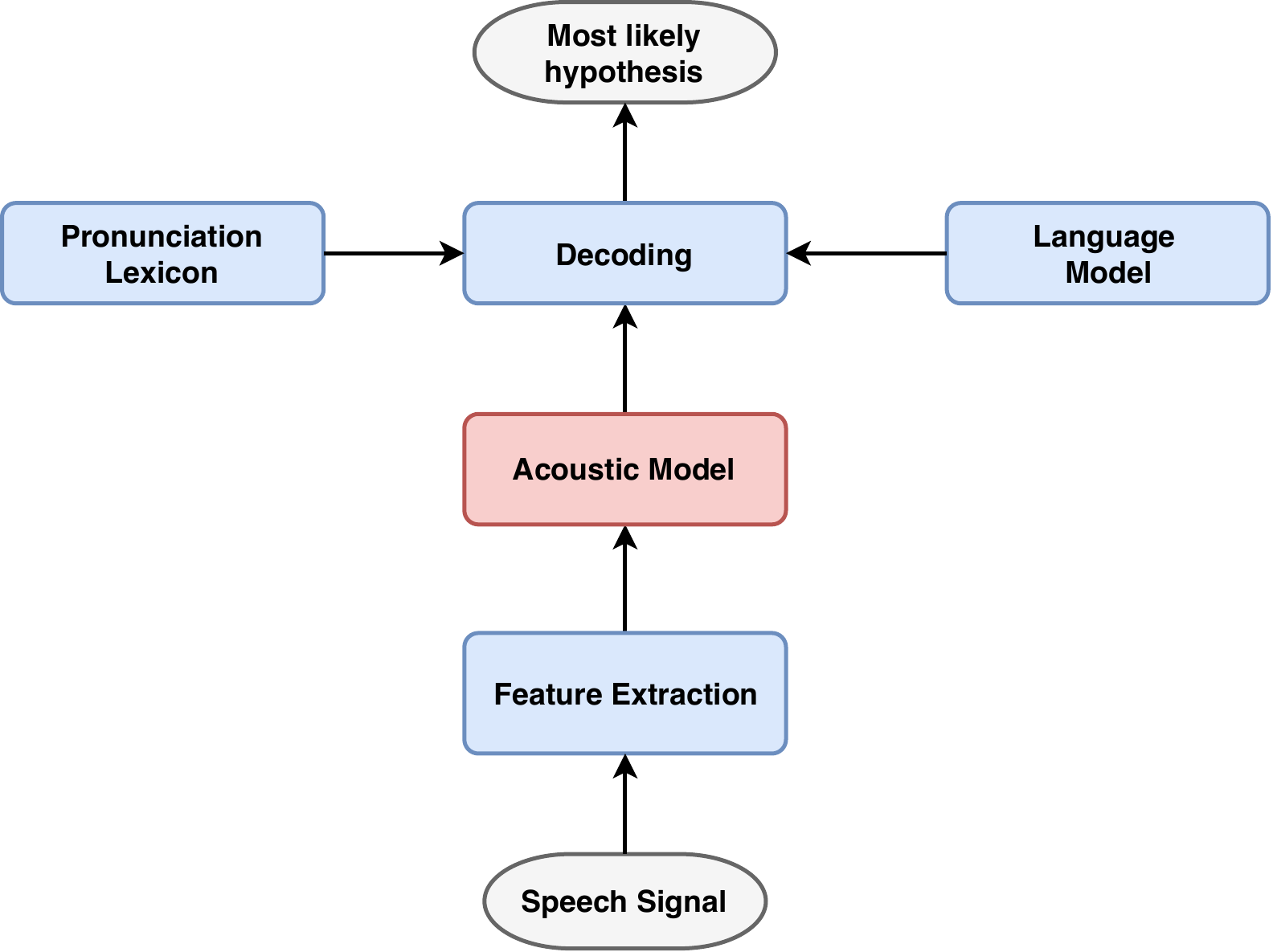}}
	\centering
	\caption{Block diagram of a conventional ASR system. The acoustic and linguistic components are incorporated to jointly determine the most likely hypothesis.}
	\label{asrblock}
\end{figure}

As shown in Figure~\ref{asrblock}, conventional ASR systems uses acoustic and linguistic information preserved in three distinct components to convert speech signals to the corresponding text: (1) an acoustic model for preserving the statistical representations of different speech units, e.g. phones, from speech features, (2) a language model for assigning probabilities to the co-occurring word sequences and (3) a pronunciation lexicon for mapping the phonetic transcriptions to orthography. These resources are jointly used to determine the most likely hypothesis in the decoding stage.

The acoustic modeling has been achieved using various statistical models such as Gaussian Mixture Models (GMM) for assigning frame-level phone posteriors in conjunction with a Hidden Markov Model (HMM) for duration modeling~\citep{YuDeng15}. More recently, ANN-based approaches have become the standard acoustic models providing state-of-the-art performance across a wide spectrum of ASR tasks~\citep{hinton2012deep}. Together with the numerous ANN architectures explored for acoustic modeling, several end-to-end ANN architectures have been proposed for directly mapping speech features to text with optional use of the other linguistic components~\citep{graves2014,chan2016,watanabe2017}.

The probabilistic definition of acoustic modeling becomes more evident via the Bayesian formulation of the speech recognition task. Given a target speech signal that segmented into T overlapped frames, the resulting frame-wise features can be represented as $\textbf{O} = [\textbf{o}_\textsubscript{1}, \textbf{o}_\textsubscript{2},..., \textbf{o}_\textsubscript{T}]$. An ASR system assigns the probability $P(\textbf{W}|\textbf{O})$ to all possible word sequences $\textbf{W} = [w_\textsubscript{1}, w_\textsubscript{2},...]$, and the word sequence $\hat{\textbf{W}}$ with the highest probability is the recognized output,
\begin{equation}
 \hat{\textbf{W}} = \argmax_{\textbf{W}} P(\textbf{W}|\textbf{O})
\end{equation}
The probability $P(\textbf{W}|\textbf{O})$ can be decomposed into two parts by applying the Bayes' rule as below,
\begin{equation}
 \hat{\textbf{W}} = \argmax_{\textbf{W}} \frac{P(\textbf{O}|\textbf{W})P(\textbf{W})}{P(\textbf{O})} 
\end{equation}
$P(\textbf{O})$ can be omitted as it does not depend on $\textbf{W}$. This results in
\begin{equation}
\label{eq:search}
\hat{\textbf{W}} = \argmax_{\textbf{W}} P(\textbf{O}|\textbf{W})P(\textbf{W})
\end{equation}
which formally defines the theoretical foundation that are grounded in conventional ASR systems. $P(\textbf{W})$ is the prior probability of the word sequence $\textbf{W}$ and this probability is provided by the language model which is trained on a large written corpus of the target language. $P(\textbf{O}|\textbf{W})$ is the \textit{likelihood} of the observed feature sequence $\textbf{O}$ given the word sequence $\textbf{W}$, and this probability is associated with the acoustic model. The acoustic model captures the information about the acoustic component of speech signals, aiming to classify different acoustic units accurately. Traditionally, each phone in the phonetic alphabet is modeled using multiple three-state HMM models for different preceding and following phonetic context (triphone)~\citep{lee1990}. The emission probability of these HMM states are shared (tied) among different models to reduce the number of model parameters~\citep{hwang1993}. The output layer of the ANN-based acoustic model is designed accordingly and trained to assign these frame-level tied triphone HMM state (senone) probabilities~\citep{dahl2012}. The output layer uses the softmax function to normalize the output into a probability distribution. These values are scaled with the prior probabilities of each class, obtained from the training data, to determine the likelihood values. These likelihood values are later combined with the probabilities assigned by the language model during the decoding stage so as to find the most likely hypothesis.

Speech features, used as inputs to the acoustic model, describe the spectrotemporal dynamics of the speech signal and discriminate among different phones in the target language. Mel-frequency cepstral coefficients(MFCC)~\citep{davis1980} features are commonly used in conjunction with the GMM-HMM acoustic model. The MFCC features are extracted by (1) performing short-time Fourier transform, (2) applying triangular Mel-scaled filter banks to calculate the power at each Mel frequency in log domain (FBANK) and (3) performing a discrete cosine transform to decorrelate the FBANK features. The third step is often skipped and FBANK features are often used when training ANN-based acoustic models since these models can handle correlation among features. In this work, we incorporate deep SNNs for acoustic modeling instead of the conventional ANNs and compare their ASR performance in different ASR scenarios including phone recognition, low-resourced and standard large vocabulary ASR. The ASR performance obtained using popular speech features have been reported to explore the impact of the feature representation space and its dimensionality for SNN-based acoustic models.

\subsection{Speech Recognition with Spiking Neural Network}
SNNs are well-suited for representing and processing spatial-temporal signals, they hence possess great potentials for speech recognition tasks. Tavanaei et al. \citep{ref72, ref73} proposed SNN-based feature extractors to extract discriminative features from the raw speech signal using unsupervised spiking-timing-dependent plasticity (STDP) rule. While connecting these SNN-based feature extractors with Support Vector Machine (SVM) or Hidden Markov Model (HMM) classifiers, competitive classification accuracies were demonstrated on the isolated spoken digit recognition task. Wu et al. \citep{wu2018ijcnn,wu2018spiking} introduced a SOM-SNN framework for environmental sound and speech recognition. In this framework, the biological-inspired self-organizing map (SOM) is utilized for feature representation, which maps frame-based acoustic features into a spike-based representation that is both sparse and discriminative. The temporal dynamic of the speech signal is further handled by the SNN classifier. Zhang et. al \citep{zhang2019mpd} presented a fully SNN-based speech recognition framework, wherein the spectral information of consecutive frames are encoded with threshold coding and subsequently classified by the SNN that is trained with a novel membrane potential-driven aggregate-labeling learning algorithm. 

Recurrent network of spiking neurons (RSNNs) exhibit greater memory capacity than the aforementioned feedforward frameworks. They can capture long temporal information that are useful for speech recognition tasks. In \citep{zhang2015digital}, Zhang et al. presented a spiking liquid-state machine (LSM) speech recognition framework which is attractive for low-power very-large-scale-integration (VLSI) implementation. Bellec et al. recently demonstrated state-of-the-art phone recognition accuracy on the TIMIT dataset by adding neuronal adaptation mechanism to the vanilla RSNNs \citep{bellec2018long}. It is the first time that RSNNs approaching the performance of LSTM networks \citep{greff2016lstm} on the speech recognition task. These preliminary works on the SNN-based ASR systems are however limited to the phone classification or small vocabulary isolated spoken digit recognition tasks. In this work, we apply deep SNNs to LVCSR tasks and demonstrate competitive accuracies over the comparable ANN-based ASR systems.

\section{Methods}
\label{sec:method}
\subsection{Spiking Neuron Model}
\label{subsec:neuronmodel}
\begin{figure}[htb]
	\centering
	\centerline
	{\includegraphics[width = 12 cm]{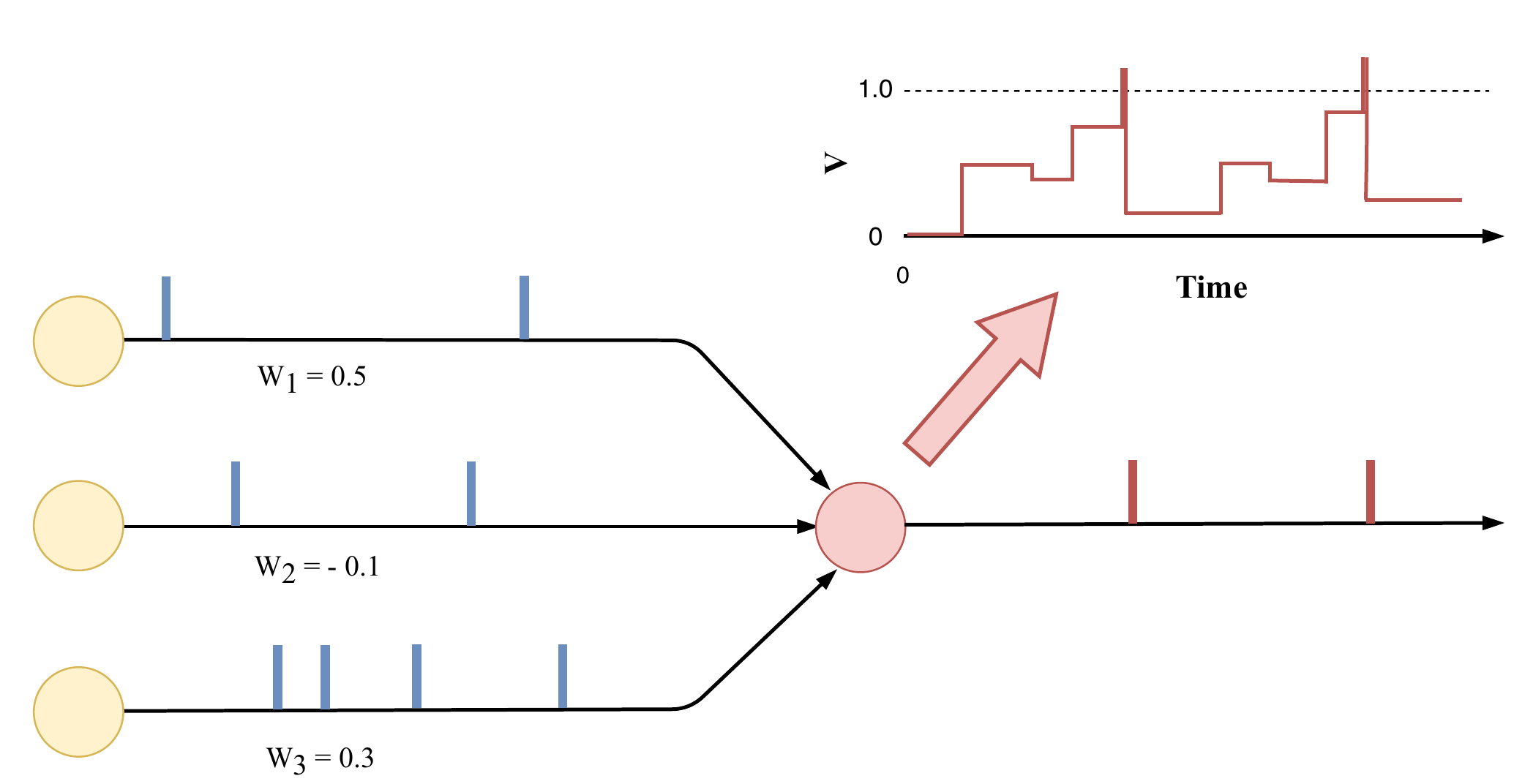}}
	\caption{The neuronal dynamic of an integrate-and-fire neuron (red). In this example, three pre-synaptic neurons are sending asynchronous spike trains to this neuron. Output spikes are generated when the membrane potential $V$ crosses the firing threshold (top right corner).}
	\label{ifNeuron}
\end{figure}

\begin{figure}[htb]
	\centering
	\centerline
	{\includegraphics[width = 10 cm]{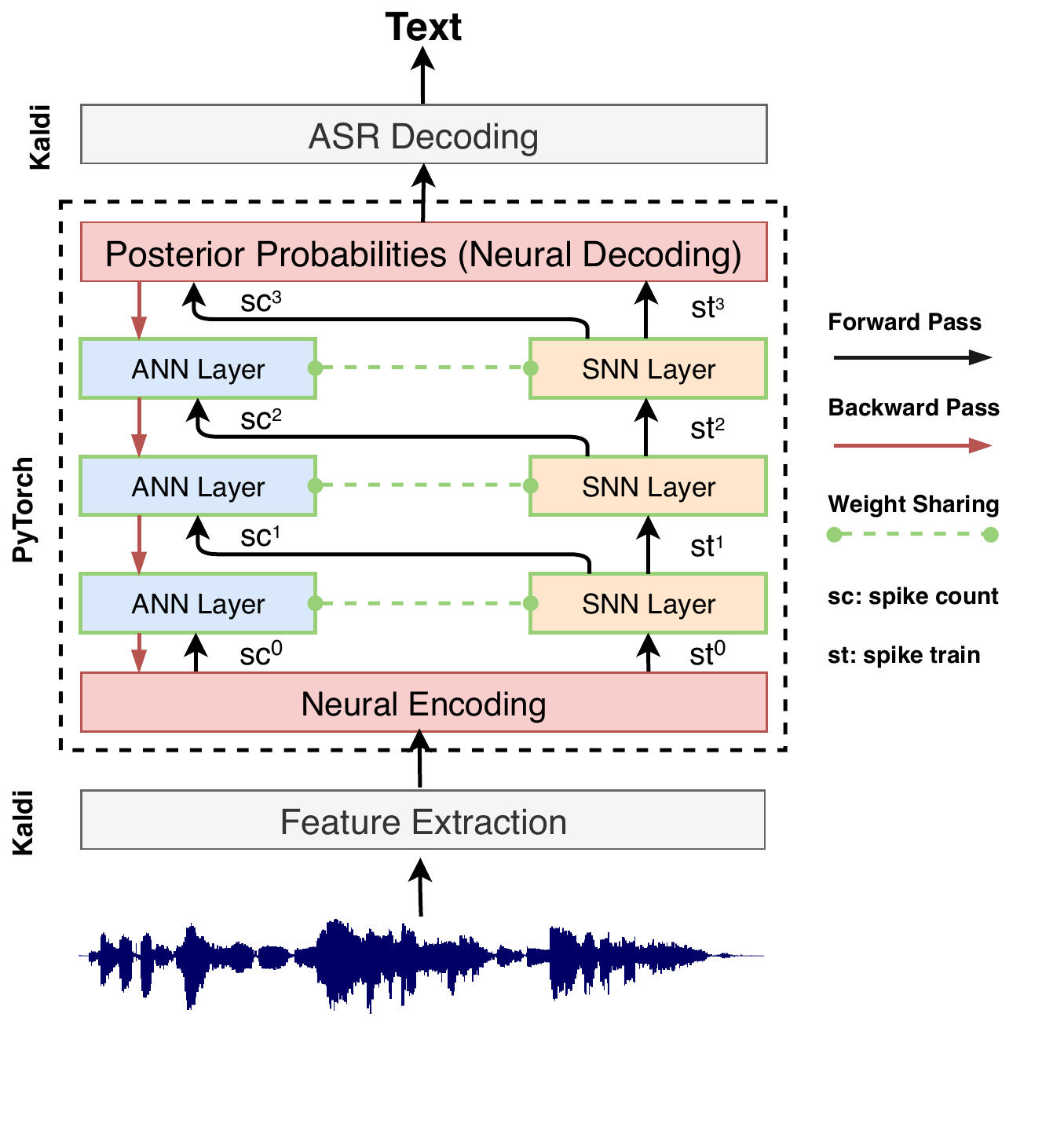}}
	\caption{System flowchart for SNN training within a tandem neural network, wherein SNN layers are used in the forward pass to determine the spike count and spike train. The ANN layers are used for error back-propagation to approximate the gradient of the coupled SNN layers.}
	\label{learningRule}
\end{figure}

As shown in Figure \ref{learningRule}, the frame-based features are extracted and input into the SNN-based acoustic models. Given the short temporal duration of segmented frames and the slow variation of speech signals, these features are typically assumed to be stationary over the short time-period of segmented frames. In this work, we use the integrate-and-fire (IF) neuron model with reset by subtraction scheme \citep{ethImageNet}, which can effectively process these stationary frame-based features with minimal computational costs. At each time step $t$ of a discrete-time simulation, the incoming spikes to neuron $j$ at layer $l$ are transduced into synaptic current as follows
\begin{equation}
z_j^l(t) = \sum\nolimits_i {w_{ji}^{l - 1} \cdot \theta _i^{l - 1}} (t) + b_j^l \slash T_{e}\\
\label{eq1}
\end{equation}
where $\theta _i^{l - 1}(t)$ indicates the occurrence of an input spike from afferent neuron $i$ at time step $t$. In addition, the $w_{ji}^{l - 1}$ denotes the synaptic weight that connects presynaptic neuron $i$ from layer $l-1$. Here, $b_j^l/T_{e}$ can be interpreted as a constant injecting current across the encoding time window of size $T_{e}$, and $b_j$ is determined from the bias term of the coupled analog neurons which will be explained in the tandem learning section. As shown in Figure \ref{ifNeuron}, neuron $j$ integrates the input current $z_j^l(t)$ into its membrane potential $V_j^l(t)$ as per Eq. \ref{eq2}. The $V_j^l(0)$ is reset and initialized to zero for every new frame-based feature input. Without loss of generality, a unitary membrane resistance is assumed here. An output spike is generated whenever $V_j^l(t)$ crosses the firing threshold $\vartheta$ (Eq. \ref{eq4}), which we set to a value of $1$ for all the experiments by assuming that all synaptic weights are normalized with respect to the $\vartheta$.
\begin{equation}
V_j^l(t) = V_j^l(t - 1) + z_j^l(t) - \vartheta \cdot \theta _j^l(t - 1)
\label{eq2}
\end{equation}
\begin{equation}
V_j^l(0) =  0
\label{eq3}
\end{equation}
\begin{equation}
\theta _j^l(t) = \Theta (V_j^l(t) - \vartheta) \;\; with \;\; \Theta (x) = \left\{\begin{array}{l}1,\;\;\; if\;x \ge 0\\
0,\;\;\;otherwise\;
\end{array} \right.
\label{eq4}
\end{equation}
According to Eqs. \ref{eq1} and \ref{eq2}, the free aggregated membrane potential of neuron $j$ (no firing) in layer $l$ can be expressed as
\begin{equation}
V_j^{l,f}= \sum\nolimits_i {w_{ji}^{l - 1} \cdot c_i^{l - 1}} + b_j^l \\
\label{eq5}
\end{equation}
where $c_i^{l - 1}$ is the input spike count from pre-synaptic neuron $i$ at layer $l-1$ as per Eq. \ref{sc}.
\begin{equation}
c_i^{l-1} = \sum\nolimits_{t=1}^{T_e} {\theta _i^{l-1}(t)}. \\
\label{sc}
\end{equation}
The $V_j^{l,f}$ summarizes the aggregate membrane potential contributions of the incoming spikes from pre-synaptic neurons while ignoring their temporal distribution. As will be explained in the tandem learning framework section, this intermediate quantity links the SNN layers to the coupled ANN layers for parameter optimization.

\subsection{Neural Coding Scheme}
SNNs process information transmitted via spike trains, therefore, special mechanisms are required to encode the continuous-valued feature vectors into spike trains and decode the classification results from the activity of output neurons. To this end, we adopt the spiking neural encoding scheme that proposed in the tandem learning framework \citep{hybridlearning}. This encoding scheme first transforms frame-based input feature vector $X^0$ (e.g., MFCC or FBANK features), where ${X^0} = {[x_1^0,x_2^0, \cdot  \cdot  \cdot ,x_n^0]^T}$, through a weighted layer of rectified linear unit (ReLU) neurons as follows 
\begin{equation}
V_j^{0,f}(0) \equiv a_j^{0} = f(\sum\nolimits_i {w_{ji}^{0} \cdot x_i^{0}} + b_j^0) \\
\label{encode_eq}
\end{equation}
where $w_{ji}^{0}$ is the strength of the synaptic connection between the input $x_i^{0}$ and ReLU neuron $j$. The $b_j^0$ is the corresponding bias term of the neuron $j$, and $f(\cdot)$ denotes the ReLU activation function. The free aggregate membrane potential $V_j^{0,f}(0)$ is defined to be equal to the activation value $a_j^{0}$ of the ReLU neuron $j$.  We distribute this quantity over the encoding time window $T_e$ and represent it with spike trains as per Eqs. \ref{encode_spike} and \ref{encode_spike2}.
\begin{equation}
\theta _j^0(t) = \Theta (V_j^{0,f}(t-1) - \vartheta)\;\; 
\label{encode_spike}
\end{equation}
\begin{equation}
V_j^{0,f}(t) = V_j^{0,f}(t-1) - \vartheta \cdot \theta _j^0(t) \;\;
\label{encode_spike2}
\end{equation}
Altogether, the spike train $s^0$ and spike count $c^0$ that output from the neural encoding layer can be represented as follows
\begin{equation}
{s^0} = \{{\theta ^0}(1),...,{\theta ^0}({T_{e}})\}
\end{equation}
\begin{equation}
{c^0} = \sum\nolimits_{t = 1}^{{T_{e}}} {{\theta ^0}(t)} 
\end{equation}
This encoding layer performs weighted transformation inside an end-to-end learning framework. It transforms the original input representation to match the size of the encoding time window $T_e$ and represents the transformed information via spike trains. This encoding scheme is beneficial for rapid inference since the input information can be effectively encoded within a short encoding window. Start from this neural encoding layer, as shown in Figure \ref{learningRule}, we input the spike count ${c^l} $ and ${s^l}$ to subsequent ANN and SNN layers for tandem learning.

To ensure smooth learning with high precision error gradients derived at the output layer, we use the free aggregate membrane potential of output spiking neurons for neural decoding. Considering that the dimensionality of input feature vectors and output classes are much smaller than that of hidden layers, the computation required will be limited when deploying these two layers onto the edge devices.

\subsection{Tandem Learning for Training Deep SNNs}
\label{sec:trainSNN}
Although IF neurons do not emulate rich temporal dynamics of biological neurons, they are however ideal for working with the neural representation that employed in this work, where spike timings play an insignificant role. It is worth noting that connections are commonly drawn between the activation value of ReLU neurons and the steady-state firing rate of IF neurons \citep{ethImageNet}. Here, we present a recently proposed SNN learning rule, under the tandem neural network configuration, that exploits such a connection between the activation value of ANN neurons and the spike count of IF neurons. 

By neglecting the temporal dynamic of IF neuron that due to the temporal distribution of incoming spike trains, we may consider the $V_j^{l,f}$ as the main information carrier for SNN layers. The following one-to-one correspondence between the free aggregate membrane potential $V_j^{l,f}$ of spiking neurons and the pseudo `spike count' $a_j^l$ of artificial neurons can be established. 
\begin{equation}
\begin{aligned}
a_j^l = min\left(\left\lfloor {V_j^{l,f}}\slash{\vartheta} \right\rfloor \cdot\left( {V_j^{l,f} > 0} \right), T_{e} \right) 
\end{aligned}
\label{eq6}
\end{equation}
As shown in Figure \ref{learningRule}, during the activation forward propagation, the SNN layers are used to determine the exact spike representation which then propagate the aggregate spike counts and spike trains to the subsequent ANN and SNN layers, respectively. This interlaced layer structure ensures the information that forward propagated to the coupled ANN and SNN layers are synchronized. Taking Eq. \ref{eq6} as the activation function of ANN layers and using straight-through estimator \citep{bengio2013estimating} to address the discontinuity of the rounding operation, we can use the error gradients derived from ANN layers to approximate those of the coupled SNN layers. It worth noting that the ANN is just an auxiliary structure to facilitates the training of SNN, while only SNN is used during inference.

Notably, competitive classification accuracies are demonstrated with this tandem learning rule for the image classification on the ImageNet dataset \citep{hybridlearning}. By analyzing the relationships between the approximated `spike count' $a_j^l$ and the actual spike count $c_j^{l}$ in a high-dimensional space, Wu et al. have argued that the modified learning dynamic of such a decoupled network can approximate that of an intact ANN. The details of this tandem learning rule are provided in the Algorithm \ref{algo}. 

\subsection{SNN-based Acoustic Modeling}
To train the deep SNN-based acoustic models, which is the main contribution of this work, several popular speech features have been extracted from the training recordings as described in Section~\ref{ssec:lvasr}. Before being fed into the SNNs, these input speech features are contextualized by splicing multiple frames so as to exploit more temporal context information. Before training the SNN-based acoustic model, alignments of the speech features with the target senone labels are obtained using a conventional GMM-HMM-based ASR system similar to that described in \citep{dahl2012}. These frame-level alignments enable the training of the deep SNN acoustic model with the tandem learning approach. During the training, the deep SNN learns to map input speech features to posterior probabilities of senones (cf. Section~\ref{ssec:lvasr}) by passing the input speech frames through multiple layers of spiking neurons.

During the inference phase, the acoustic scores provided by the trained SNN model are combined with the information stored in the language model and pronunciation lexicon. It is a common practice to use the weighted finite state transducers (WFST)~\citep{mohri2002} as a unified representation of different ASR resources for creating the search graph containing possible hypotheses. The main motivation for using the WFST-based decoding is: (1) the straightforward composition of different ASR resources for constructing a mapping from HMM states to word sequences and (2) the existence of efficient search algorithms operating on WFST that speed up the decoding process. As a result of the search process, the most likely hypotheses are found and stored in the form of a lattice. The ASR output is chosen based on the weighted sum of the acoustic and language model scores belonging to hypotheses in the lattice. For further details of the WFST-based decoding approach used in this work, we refer the reader to~\citep{povey2012}. In the following sections, we describe the ASR experiments conducted to evaluate the recognition performance of the proposed SNN-based acoustic modeling in several recognition scenarios.

\subsection{Training and Evaluation}
\label{sec:exp}
\subsubsection{Datasets}
The performance of the proposed SNN-based acoustic models is investigated in three different ASR tasks: (1) phone recognition using the TIMIT corpus~\citep{timit}, (2) low-resourced ASR task using the FAME code-switching Frisian-Dutch corpus~\citep{yilmaz2016} and (3) standard large-vocabulary continuous ASR task using the Librispeech corpus~\citep{librispeech}. All speech data used in the experiments has a sampling frequency of 16 kHz.

The train, development and test sets of the standard TIMIT corpus contain 3,696, 400 and 192 utterances from 462, 50 and 24 speakers, respectively. Each utterance is phonetically transcribed using a phonetic alphabet consisting of 48 phones in total. The training data of the FAME corpus comprises of 8.5 hours and 3 hours of broadcast speech from Frisian and Dutch speakers, respectively. The training utterances are spoken by 382 speakers in total. This bilingual dataset contains Frisian-only and Dutch-only utterances as well as mixed utterances with inter-sentential, intra-sentential and intra-word code-switching~\citep{myers1989}. The development and test sets consist of 1 hour of speech from Frisian speakers and 20 minutes of speech from Dutch speakers each. The total number of speakers is 61 in the development set and 54 in the test set.

The Librispeech corpus contains 1,000 hours of reading speech in total collected from audiobooks. This publicly available corpus\footnote{www.openslr.org/resources/12} has been considered as a popular benchmark for ASR algorithms with multiple training and testing settings. In the ASR experiments, we train acoustic models using the 100 (train\_clean\_100) and 360 (train\_clean\_360) hours of speech and apply these models to the clean development (dev\_clean) and test (test\_clean) sets. Further details about this corpus can be found in~\citep{librispeech}.

\subsubsection{Implementation Details}
All ASR experiments are performed using the PyTorch-Kaldi ASR toolkit~\citep{pytorch-kaldi}. This recently introduced toolkit inherits the flexibility of PyTorch toolkit~\citep{paszke2017automatic} for ANN-based acoustic model development and the efficiency of Kaldi ASR toolkit~\citep{povey2011kaldi}. We implement the SNN tandem learning rule in PyTorch and integrate it into the PyTorch-Kaldi toolkit for training the proposed SNN-based acoustic models (cf. Figure~\ref{learningRule}). The PyTorch implementation of the described SNN acoustic models will be made available online soon. For the baseline ANN models, the standard multi-layer perceptron recipes are used. The Kaldi toolkit is used for obtaining the initial alignments, feature extraction, graph creation, and decoding.

For all recognition scenarios, ANNs and SNNs are constructed with 5 hidden layers and 2048 hidden units each using the ReLU activation function. Each fully-connected layer is followed by a batch normalization layer and a dropout layer with a drop probability of 10\% to prevent overfitting. We train these models using various popular speech features including the 13-dimensional Mel-frequency cepstral coefficient (MFCC) feature, 23-dimensional Mel-filterbank (FBANK) feature, and higher resolution 40-dimensional MFCC and FBANK features. We further extract feature space maximum likelihood linear regression (FMLLR)~\citep{gales1998} features to explore the impact of speaker-dependent features. All features include the deltas and delta-deltas; mean and variance normalization are applied before the splicing. The time context size is set to 11 frames by concatenating 5 frames preceding and following. All features are encoded within a short time window of 10-time steps for SNN simulations.

The neural network training is performed by mini-batch Stochastic Gradient Descent (SGD) with an initial learning rate of 0.08 and a minibatch size of 128.~The learning rate is halved if the improvement is less than a preset threshold of 0.001.~The final acoustic models of the TIMIT and FAME corpora are obtained after 24 training epochs, while the models of the Librispeech corpus are trained for 12 epochs.

For the TIMIT and Librispeech ASR tasks, we follow the same language model (LM) and pronunciation lexicon preparation pipeline as provided in the corresponding Kaldi recipes\footnote{https://github.com/kaldi-asr/kaldi/tree/master/egs/\{timit,librispeech\}}. The smallest 3-gram LM (\texttt{tgsmall}) of the Librispeech corpus is used to create the graph for the decoding stage. The details of the LM and lexicon used in the FAME recognition task are given in~\citep{yilmaz2018_1}. 

\subsubsection{Evaluation Metrics}
\paragraph{ASR Performance}
The phone recognition on the TIMIT corpus is reported in terms of the phone error rate (PER). The word recognition accuracies on the FAME and Librispeech corpora are reported in terms of word error rate (WER). Both metrics are calculated as the ratio of all recognition errors (insertion, deletion, and substitution) and the total number of phones or words in the reference transcriptions.

\paragraph{Energy Efficiency: Counting Synaptic Operations}
To compare the energy efficiency of ANN and its equivalent SNN implementation, we follow the convention from NC community and compute the total synaptic operations $SynOps$ that required to perform a certain task \citep{merolla2014million,ethImageNet,sengupta2019going}. For ANN, the total synaptic operations (Multiply-and-Accumulate (MAC)) per classification is defined as follows
\begin{equation}
SynOps = \sum\limits_{l = 1}^L {{f_{in}^l}}  \cdot {N_l}
\end{equation}
where $f_{in}^l$ denotes the number of fan-in connections to each neuron in layer $l$, and $N_l$ refers to the number of neurons in layer $l$. In addition, $L$ denotes the total number of network layers. Hence, given a particular network configuration, the total synaptic operations required per classification is a constant number that jointly determined by $f_{in}^l$ and ${N_l}$.

While for SNN, as per Eq. \ref{snnSynOps}, the total synaptic operations (Accumulate (AC)) required per classification are correlated with the spiking neurons' firing rate, the number of fan-out connections $f_{out}$ to neurons in the subsequent layer as well as the simulation time window $T$.
\begin{equation}
SynOps = \sum\limits_{t = 1}^T {\sum\limits_{l = 1}^{L-1} {\sum\limits_{j = 1}^{N_l} {f_{out,j}^l} } }  \cdot s_j^l(t)
\label{snnSynOps}
\end{equation}
where $s_j^l(t)$ indicates whether a spike is generated by neuron $j$ of layer $l$ at time instant $t$. 

\section{Results}
\label{sec:result}
\subsection{Phone Recognition on TIMIT Corpus}
We report the PER on the development and test sets of TIMIT corpus in Table~\ref{tab:timit}, with numbers in bold being the best performance given by the speaker-independent features. ASR performances of other state-of-the-art systems using various ANN and SNN architectures are given in the upper panel for reference purposes. As the results shown in Table~\ref{tab:timit}, the proposed SNN-based acoustic models are applicable to different speech features and provide comparable or slightly worse ASR performance than the ANNs with the same network structure. In particular, the ANN system trained with the standard 13-dimensional FBANK feature achieves the best PER of 16.9\% (18.5\%) on the development (test) set. The equivalent SNN system using the same feature achieves slightly worse PER of 17.3\% (18.7\%) on the development (test) set. Although the state-of-the-art ASR systems \citep{ravanelli2018} give approximately 1\% lower PER than the proposed SNN-based phone recognition system, it is largely credit to the longer time context explored by the recurrent Li-GRU model. 

It worth mentioning that phone recognition is still a challenging task for spiking neural networks. To the best of our knowledge, only one recent work with recurrent spiking neural networks \citep{Bellec738385} demonstrates some promising test results on this corpus with a PER of 26.4\%. In contrast, our system has achieved significantly lower PER compared to this preliminary study of SNN-based acoustic modeling. However, these results are not directly comparable since the proposed system incorporates both an acoustic and a language model during decoding unlike the system described in~\citep{Bellec738385}.

The experimental results on the TIMIT phone recognition task can be considered as an initial indicator of the compelling prospects of the SNN-based acoustic modeling. Given that the phone recognition task on TIMIT corpus is simplistic compared to the modern LVCSR tasks, we further compare the ANN and SNN performance on newer corpora designed for LVCSR experiments. 

\subsection{Low-resourced ASR on FAME Corpus}
In this section, we apply the SNN-based ASR systems to the low-resourced ASR scenario. As summarized in Table~\ref{tab:fame}, the word recognition results on the FAME corpus are reported separately for monolingual Frisian (fy), monolingual Dutch (nl) and code-switched (cs) utterances. The overall performance (all) is also included in the rightmost column. Given that 8.5 hours Frisian and 3 hours of Dutch speech is used during the training phase, we can compare the ASR performance on different subsets, i.e. fy, nl and cs, to identify the variations in the ASR performance for different levels of low-resourcedness. We omit the results on the development set as they follow a similar pattern to the results on the test set.

In this scenario, the SNN acoustic models consistently provide lower WERs than the ANN models for all speech features. Systems with the FBANK features provide lower WERs than those using MFCC features, which is in line with our observations on the TIMIT corpus. The best performance on the test set is obtained using SNN models trained on 40-dimensional FBANK features with an overall WER of 36.9\%. In contrast, the ANN model provides a WER of 39.0\% for the same setting, which is relatively 5.4\% worse than the SNN model. Moreover, the SNN-based acoustic models achieve a relative improvement of 4.7\%, 5.2\% and 8.2\% on the fy, nl and cs subsets of the test set, respectively. These steady improvements in the recognition accuracies highlight the effectiveness of the SNN-based acoustic modeling in scenarios with limited training data compared to the conventional ANN models. The improved ASR performance with SNNs, in the low-resourced setting, may credit to the noisy weight updates derived by the coupled ANN layers of the tandem learning framework \citep{hybridlearning}. It has been recognized that introducing noises into the training stage improves the generalization capability of ANN-based ASR systems~\citep{yin2015noisy}. As a result, the noisy training of the tandem learning is expected to improve the recognition performance in low-resourced scenarios. Further investigation on the impact of this noisy training procedure remains as future work.

\subsection{LVCSR experiments on Librispeech Corpus}
In the final set of ASR experiments, we train acoustic models using the official 100-hour and 360-hour training subsets of the Librispeech corpus to compare the recognition performance of ANN and SNN models in a standard LVCSR scenario. As the results given in the middle panel of Table~\ref{tab:libri}, for 100 hours of training data, the ANN systems perform marginally better than the corresponding SNN systems across all different speech features. The absolute WER differences range from 0.1\% to 0.6\%. These marginal performance degradations of the SNN models is likely due to the reduced representation power of using discrete spike counts. Nevertheless, these results are promising even when comparing to the state-of-the-art ASR systems using more complex ANN architectures as provided in the upper panel of Table~\ref{tab:libri}. 

It worth noting that both ANN and SNN systems can take benefit of an increased amount of training data. When increasing the training data from 100 hours to 360 hours, the WERs of the best SNN models reduced from 10.0\% (10.3\%) to 9.2\% (9.4\%) for the development (test) sets, respectively. To the best of our knowledge, it is the first time that SNN-based acoustic models have achieved comparable results over the ANN models for LVCSR tasks. These results suggest that SNNs are potentially good candidates for acoustic modeling. 

\subsection{Energy Efficiency of SNN-based ASR Systems}
\begin{figure}[htb]
	\centering
	\centerline
	{\includegraphics[width=12cm]{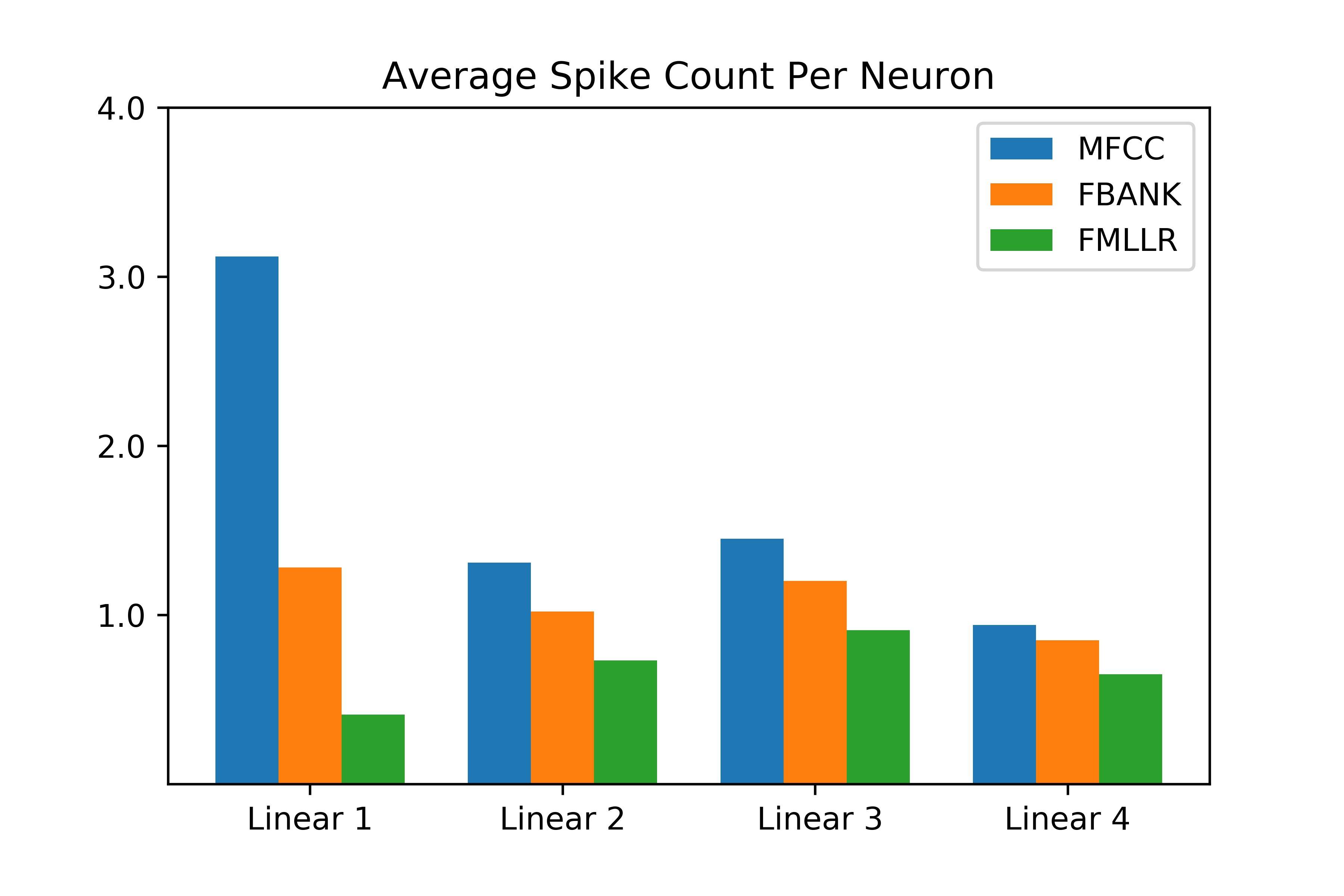}}
	\caption{Average spike count per neuron of different SNN layers on the TIMIT corpus. The results of different input features are color-coded. Sparse neuronal activities can be observed in this bar chart.} %suggesting low computational costs are required for the SNN.}
	\label{spikeRate}
\end{figure}

In addition to the promising modeling capability, the SNN-based ASR systems can achieve unprecedented performance gain when implemented on the low-power neuromorphic chips. In this section, we shed light on this prospect by comparing the energy efficiency of ANN- and SNN-based acoustic models. Given that data movements are the most energy-consuming operations for data-driven AI applications, we calculate the average synaptic operations on 5 randomly chosen utterances from the TIMIT corpus and report the ratio of average synaptic operations required per feature classification (SynOps(SNN)\,/\,SynOps(ANN)). To investigate the effect of different feature representations, we repeat our analysis on the 40-dimensional MFCC, FBANK and FMLLR features as summarized in Table~\ref{synops} and Figure~\ref{spikeRate}.

Taking advantage of the short encoding time window ($T_e=10$), the sparse neuronal activities are observed for all network layers as shown in Figure~\ref{spikeRate}. Among the three features explored in this experiment, it is interesting to note the FMLLR feature achieves the lowest average spike rate. It is likely due to the more discriminative nature of the speaker-dependent feature, while it worth to note that the FMLLR feature is not always available in all ASR scenarios. As provided in Table~\ref{synops}, the SNN implementations taking MFCC, FBANK and FMLLR input features require 1.72, 1.10 and 0.68 times synaptic operations to their ANN counterparts, respectively. Although the average number of synaptic operations required for SNNs that using MFCC and FBANK features are slightly higher than the ANNs, the AC operations performed on SNNs are much cheaper than the MAC operations required for ANNs. One recent study on the Global Foundry 28 nm process has revealed that MAC operations are 14 times more costly than AC operations and requires 21 times more chip area~\citep{ethImageNet}. Therefore, when deploying SNNs onto the emerging neuromorphic chips for inference~\citep{merolla2014million, davies2018loihi}, we expect to receive at least an order of magnitude energy and chip area savings. While the actual energy savings for SNN-based acoustic models are dependent on the chip architectures and materials used, which is outside the scope of this work. 

\section{Discussion}
\label{sec:dis}
The remarkable progress in the automatic speech recognition systems has revolutionized the human-computer interface. The rapid growing demands of ASR services have raised concerns on computational efficiency, real-time performance, and data security, etc. It, therefore, motivates novel solutions to address all those concerns. As inspired by the event-driven computation that observed in the biological neural systems, we explore using brain-inspired spiking neural networks for large vocabulary ASR tasks. For this purpose, we proposed a novel SNN-based ASR framework, wherein the SNN is used for acoustic modeling and map the frame-level features into a set of acoustic units. These frame-level outputs will further integrate the word-level information from the corresponding language model to find the most likely word sequence corresponding to the input speech signal. 

\subsection{Superior Speech Recognition Performance with SNNs}
The phone and word recognition experiments on the well-known TIMIT and Librispeech benchmarks have demonstrated the promising modeling capacity of SNN acoustic models and their applicability to different input features. These preliminary results have shown that the recognition performance of SNNs is either comparable or slightly worse than the ANNs with the same network architecture on the TIMIT and Librispeech benchmarks. A possible reason for this performance degradation is the reduced representation power of the discrete neural representation (i.e., spike counts) as compared to the continuous floating-point representation of the ANNs \citep{hybridlearning}. This performance gap could potentially be closed by extending the encoding window $T_e$ of SNNs. Moreover, the recognition performance of ANN and SNN models in a low-resourced scenario is also investigated. In this scenario, the SNN acoustic models outperform the conventional ANNs that could be attributed to the noisy training of the tandem learning framework, wherein error gradients of the SNN layers are approximated from the coupled ANN layers. 

The neural encoding scheme adopted in this work allows input features to be encoded inside a short encoding time window for rapid processing by SNNs. It is attractive for the time-synchronous ASR tasks that require real-time performance. The preliminary study of the energy efficiency on the TIMIT corpus reveals at least an order of magnitude energy and chip area savings, as compared to the equivalent ANNs, can be achieved when deploying the offline trained SNNs onto neuromorphic chips. The recent study of a keyword spotting task on the Loihi neuromorphic research chip \citep{blouw2019benchmarking} has also demonstrated the compelling energy savings, real-time performance and good scalability of emerging NC architectures over conventional low-power AI chips designed for ANNs. 

\subsection{Development of SNN-based ASR Systems}
The active development of open-source software toolkits plays a significant role in the rapid progress of ASR research, instances include the Kaldi \citep{povey2011kaldi} and ESPnet \citep{watanabe2018espnet}. In this work, we demonstrate that state-of-the-art SNN acoustic models can be easily developed in PyTorch and integrated into the PyTorch-Kaldi Speech Recognition Toolkit \citep{pytorch-kaldi}. This software toolkit integrates the efficiency of Kaldi and the flexibility of PyTorch, therefore, it can support the rapid development of SNN-based ASR systems. 

\subsection{Future Directions}
The recurrent neural networks have shown great modeling capability for temporal signals by exploring long temporal context information in the input signals \citep{graves2014towards}. As future work, we will explore the recurrent networks of spiking neurons for large-vocabulary ASR tasks to further improve the recognition performance.

The substantial research efforts are devoted to reducing the computational cost and memory footprint of ANNs during inference, instances include network compression \citep{han2015deep}, network quantization \citep{courbariaux2016binarized, zhou2016dorefa} and knowledge distillation \citep{hinton2015distilling}. While the computational paradigm underlying the efficient biological neural networks is fundamentally different from ANNs and hence fosters enormous potentials for neuromorphic computing architectures. Furthermore, grounded on the same connectionism principle, the information of both ANN and SNN are encoded in the network connectivity and connection strength. Therefore, SNN can also take benefits from these early research works on the network compression and quantization of ANNs to further reduce its memory footprint and computation cost \citep{deng2019comprehensive}.

The event-driven silicon cochlea audio sensors \citep{ref45} are designed to mimic the functional mechanism of human cochlea and transform input audio signals into spiking events. Given temporally sparse information is transmitted in the surrounding environment, these sensors have shown greater coding efficiency than conventional microphone sensors \citep{liu2019event}. There are some interesting preliminary ASR studies explore the input spiking events captured by these silicon cochlea sensors \citep{ref69, acharya2018comparison}. However, the scale of the ASR tasks explored in these studies is relatively small comparing to modern ASR benchmarks due to the limited availability of event-based ASR corpora. Pan et al. \citep{pan2019efficient} recently proposed an efficient and perceptually motivated auditory neural encoding scheme to encode the large-scale ASR corpora collected by microphone sensors into spiking events. With this encoding scheme, approximately 50\% spiking events can be reduced with negligible interference to the perceptual quality of inputs audio signals. Taking benefits from these earlier research on the neuromorphic auditory front-end, we are expecting to further improve the energy efficiency of SNN-based ASR systems.

The promising initial results demonstrated by the SNN-based large vocabulary ASR systems in this work is the first step towards a myriad  opportunities for the integration of state-of-the-art ASR engines into mobile and embedded devices with power restrictions. In the long run, the SNN-based ASR systems are expected to take benefits from ever-growing research on novel neuromoprhic auditory front-end, SNN architectures, neuromorphic computing architectures and ultra-low-power non-volatile memory devices to further improve the computing performance.

%\subsection{Heading Levels}
%
%%There are 5 heading levels
%\subsection{Level 2}
%\subsubsection{Level 3}
%\paragraph{Level 4}
%\subparagraph{Level 5}

\section*{Conflict of Interest Statement} 
The authors declare that the research was conducted in the absence of any commercial or financial relationships that could be construed as a potential conflict of interest.

\section*{Author Contributions}
JBW and EY designed and conducted all the experiments. All authors contributed to the results interpretation and writing. 

\section*{Funding}
This research is supported by Programmatic Grant No. A1687b0033 from the Singapore Government's Research, Innovation and Enterprise 2020 plan (Advanced Manufacturing and Engineering domain). JBW is also partially supported by the Zhejiang Lab’s International Talent Fund for Young Professionals.

%\section*{Acknowledgments}
%This is a short text to acknowledge the contributions of specific colleagues, institutions, or agencies that aided the efforts of the authors.

\bibliographystyle{frontiersinSCNS_ENG_HUMS} % for Science, Engineering and Humanities and Social Sciences articles, for
\bibliography{strings_EY_JW_v6}

\pagebreak
\section*{Algorithm}
\begin{algorithm}	
	\DontPrintSemicolon
	\KwInput{Input frame-based feature vectors $X^0$, target label $Y$, network parameters $w$, neural encoding window size $T_{e}$}
	\KwOutput{Updated network parameters $w$}
	\vspace{4mm}
	\KwFw{}       % Set the Forward Pass
	${c}^0, {s}^0$ = Neural\_Encoding($X^0$) \\
	\For{layer $l$ = 1 to N-1}
	{
		\tcp*[l]{\scriptsize State Update of the ANN Layer} 
		$a^l$ = ANN.layer[$l$].forward($c^{l-1}$, $w^{l - 1}$) $^{*}$ \\
		\For{t = 1 to $T_{e}$}
		{\tcp*[l]{\scriptsize State Update of the SNN Layer} 
			$s^{l}[t]$ = SNN.layer[l].forward($s^{l-1}[t]$, $w^{l - 1}$)  
		}
		\tcp*[l]{\scriptsize Update the Spike Count}
		$c^{l}$ =  $\sum\nolimits_{t=1}^{T_{e}} {s^{l}[t]}$ 
	} 
	\tcc{\scriptsize Neural Decoding with the Aggregate Membrane Potential}
	$output$ = ANN.layer[N].forward($c^{N-1}$, $w^{N - 1}$)\\
	\vspace{2mm}
	\KwLoss{$E$ = LossFunction($Y, output$)} 
	\vspace{4mm}
	
	\KwBw{}
	$\frac{{\partial E}}{{\partial a^{N}}}$ = LossGradient($Y, output$) \\
	\For{layer $l$ = N-1 to 1}
	{	
		\tcp*[l]{\footnotesize Gradient Update through the ANN Layer} 
		$\frac{{\partial E}}{{\partial {a^{l - 1}}}}$ , $\frac{{\partial E}}{{\partial {w^{l - 1}}}}$ = ANN.layer[$l$].backward($\frac{{\partial E}}{{\partial {a^{l}}}}$ , $c^{l-1}$, $w^{l - 1}$) 
	}
	\vspace{2mm}
	Update parameters of the ANN layer based on the calculated gradients. \\
	Copy the updated parameters to the corresponding SNN layer. \\
	\vspace{2mm}
	\KwNote{}
	$^{*}$ For inference, state updates are performed on the SNN layers entirely.
	\caption{Pseudo Codes For The Tandem Learning Rule}
	\label{algo}
\end{algorithm}

\pagebreak
\section*{Tables}
\begin{table}[!htp]
	\centering
	%\addtolength{\tabcolsep}{-3pt}
	\caption{PER (\%) on the TIMIT development and test sets. The upper panel reports the results of various ANN and SNN architectures from the literatures, and the lower panel presents the results achieved by the ANN and SNN models in this work (AM: acoustic model, *: the best result to date).}
	\vspace{0.25cm}
	\resizebox{14 cm}{!}{
		\tiny
		\begin{tabular}{| l | c | c c | c c |}
			\hline
			\multicolumn{2}{|c|}{} & \multicolumn{4}{c|}{Test} \\
			\hline\hline
			\multicolumn{2}{|c|}{\diagbox{Features}{AM}}  & \multicolumn{2}{c|}{Li-GRU\citep{ravanelli2018}}&
			\multicolumn{2}{c|}{RSNN\citep{Bellec738385}} \\
			\hline
			\multicolumn{2}{|l|}{MFCC (13-dim.)}    & \multicolumn{2}{c|}{16.7} & \multicolumn{2}{c|}{26.4} \\
			\hline
			\multicolumn{2}{|l|}{FBANK (13-dim.)}   & \multicolumn{2}{c|}{15.8} & \multicolumn{2}{c|}{-}\\
			\hline
			\multicolumn{2}{|l|}{FMLLR}   & \multicolumn{2}{c|}{\,\,14.9*} & \multicolumn{2}{c|}{-}\\
			\hline\hline
			\multicolumn{2}{|c|}{} & \multicolumn{2}{c|}{Dev} & \multicolumn{2}{c|}{Test} \\
			\hline\hline
			\multicolumn{2}{|c|}{\diagbox{Features}{AM}}     & ANN  & SNN  & ANN  & SNN    \\
			\hline\hline
			\multicolumn{2}{|l|}{MFCC (13-dim.)}                & 17.1 & 17.8 & 18.5 & 19.1   \\
			\hline
			\multicolumn{2}{|l|}{FBANK (13-dim.)}               & \bf{16.9} & \bf{17.3} & 18.5 & \bf{18.7}   \\
			\hline
			\multicolumn{2}{|l|}{MFCC (40-dim.)}      & 17.3 & 18.2 & 18.7 & 19.8   \\
			\hline
			\multicolumn{2}{|l|}{FBANK (40-dim.)}     & \bf{16.9} & 17.8 & \bf{17.9} & 19.1   \\
			\hline\hline
			\multicolumn{2}{|l|}{FMLLR}               & 15.8 & 16.5 & 17.2 & 17.4   \\
			\hline
		\end{tabular}
	}
	\label{tab:timit}
\end{table}

\begin{table}[!htp]
	\centering
	\caption{WERs (\%) achieved on the monolingual and mixed segments of the FAME test set. The upper panel summarizes the number of words from each language subset. The middle panel provides the results of state-of-the-art ANN achitectures~\citep{yilmaz2016_2,yilmaz2018_1} for reference purposes and the lower panel presents the results achieved by the ANN and SNN models in this work (AM: acoustic model).}
	\vspace{0.25cm}
	% \addtolength{\tabcolsep}{-1pt}
	\resizebox{15 cm}{!}{
		\tiny
		\begin{tabular}{| l | p{1cm} | p{1cm} | c c c c |}
			\hline
			\multicolumn{3}{|c|}{} & fy & nl & cs & all \\
			\hline
			\multicolumn{3}{|c|}{\# of Frisian words} &  10753 & 0    & 1798 & 12551 \\
			\hline
			\multicolumn{3}{|c|}{\# of Dutch words}   &  0       & 3475 & 306  & 3781 \\
			\hline
			Speech features & \multicolumn{2}{c|}{AM} & & & & \\
			\hline\hline
			FBANK (40-dim.) & \multicolumn{2}{c|}{Kaldi-ANN\citep{yilmaz2016_2}} & 32.4 & 39.7 & 49.9 & 36.2  \\
			\hline
			MFCC (40-dim.) & \multicolumn{2}{c|}{TDNN-LSTM\citep{yilmaz2018_1}} & 31.5 & 39.5 & 47.9 & 35.2 \\
			\hline \hline
			MFCC (13-dim.)& \multicolumn{2}{c|}{ANN}  &  34.6 & 50.0 & 49.9 & 39.9 \\
			\hline
			MFCC (13-dim.)& \multicolumn{2}{c|}{SNN}  &  33.8 & 45.3 & 47.9 & 38.2 \\
			\hline
			FBANK (13-dim.) & \multicolumn{2}{c|}{ANN} &  34.3 & 47.5 & 48.1 & 39.0 \\
			\hline
			FBANK (13-dim.)& \multicolumn{2}{c|}{SNN} &  33.1 & 44.3 & 46.5 & 37.3 \\
			\hline
			MFCC (40-dim.) & \multicolumn{2}{c|}{ANN}  &  35.2 & 48.4 & 51.7 & 40.2  \\
			\hline
			MFCC (40-dim.) & \multicolumn{2}{c|}{SNN}  &  33.7 & 44.2 & 46.9 & 37.7  \\
			\hline
			FBANK (40-dim.)& \multicolumn{2}{c|}{ANN}  &  34.4 & 46.3 & 49.8 & 39.0  \\
			\hline
			FBANK (40-dim.)& \multicolumn{2}{c|}{SNN}  &  \bf{32.8} & \bf{43.9} & \bf{45.7} & \bf{36.9}  \\
			\hline
			FMLLR & \multicolumn{2}{c|}{ANN}  &  31.2 & 42.1 & 47.2 & 35.7 \\
			\hline
			FMLLR & \multicolumn{2}{c|}{SNN}  &  31.5 & 39.5 & 46.6 & 35.2 \\
			\hline
		\end{tabular}
	}
	\label{tab:fame}
\end{table}

\begin{table}[!htp]
	\centering
	\caption{WER (\%) achieved on the Librispeech development and test sets. The upper panel gives the results, with 100-hour of training data, reported at the Github repo of Kaldi and PyTorch-Kaldi. The middle and lower panel present the results achieved by ANN and SNN models in this work using 100-hour and 360-hour of training data, respectively.~(AM: acoustic model,\textsuperscript{\dag}: reported at Github repo)}
	\vspace{0.25cm}
	\resizebox{10 cm}{!}{
		\tiny
		\begin{tabular}{| l | c | c c | c c |}
			\hline
			\multicolumn{2}{|l|}{} & \multicolumn{4}{c|}{\bf{Train - 100h}} \\
			\hline
			\multicolumn{2}{|c|}{} & \multicolumn{2}{c|}{Dev} & \multicolumn{2}{c|}{Test} \\
			\hline
			& AM  & \multicolumn{2}{c|}{} & \multicolumn{2}{c|}{} \\
			\hline
			Kaldi\textsuperscript{\dag} & p-norm ANN & \multicolumn{2}{c|}{9.2} & \multicolumn{2}{c|}{9.7}\\
			\hline
			PyTorch-Kaldi\textsuperscript{\dag} & Li-GRU & \multicolumn{2}{c|}{-} & \multicolumn{2}{c|}{8.6}\\
			\hline\hline
			\multicolumn{2}{|l|}{\diagbox{Features}{AM}}    & ANN  & SNN  & ANN  & SNN  \\
			\hline\hline
			\multicolumn{2}{|l|}{MFCC}                & 10.3 & 10.5 & 10.6 & 10.9 \\
			\hline
			\multicolumn{2}{|l|}{FBANK}               & 9.6  & \bf{10.0} & 10.2 & 10.6 \\
			\hline
			\multicolumn{2}{|l|}{MFCC (40-dim.)}      & \bf{9.5}  & 10.1 & \bf{10.0} & 10.6 \\
			\hline
			\multicolumn{2}{|l|}{FBANK (40-dim.)}     & 9.6  & 10.2 & 10.1 & \bf{10.3} \\
			\hline\hline
			\multicolumn{2}{|l|}{FMLLR}               & 9.2  & 9.3  &  9.7 & 9.9  \\
			\hline\hline
			\multicolumn{2}{|c|}{} & \multicolumn{4}{c|}{\bf{Train - 360h}} \\
			\hline
			\multicolumn{2}{|c|}{} & \multicolumn{2}{c|}{Dev} & \multicolumn{2}{c|}{Test} \\
			\hline
			\multicolumn{2}{|l|}{\diagbox{Features}{AM}} & ANN  & SNN  & ANN  & SNN  \\
			\hline\hline
			\multicolumn{2}{|l|}{MFCC}                &   9.2   &  9.9 & 9.6  & 10.3 \\
			\hline
			\multicolumn{2}{|l|}{FBANK}               &   8.6   &  9.7 & 9.1  & 10.0 \\
			\hline
			\multicolumn{2}{|l|}{MFCC (40-dim.)}      &  8.6 & \bf{9.2} & \bf{8.9} & \bf{9.4} \\
			\hline
			\multicolumn{2}{|l|}{FBANK (40-dim.)}     &   \bf{8.5} &  9.4  & \bf{8.9} &  9.7 \\
			\hline\hline
			\multicolumn{2}{|l|}{FMLLR}               &   8.4   &  9.2  & 8.8  & 9.7     \\
			\hline
		\end{tabular}
	}
	\label{tab:libri}
\end{table}

\begin{table}[!htp]
	\centering
	\caption{Comparison of the computational costs between SNN and ANN. The ratio of their required total synaptic operations (SynOps(SNN)\,/\,SynOps(ANN)) is reported. It worth mentioning that ANNs use more costly MAC operations than the AC operations used in the SNNs.}
	\vspace{0.25cm}
	\resizebox{12 cm}{!}{
		\tiny
		\begin{tabular}{| l || c  c  c  c  c || c |}
			\hline
			Utterance Index     & 1   & 2    & 3   & 4   & 5   & \textbf{Avg. SynOps Ratio} \\
			\hline
			Num. of frames      & 474 & 287  & 274 & 268 & 223 &             \\
			\hline\hline
			MFCC (40-dim.)      & 1.71 & 1.73  & 1.76 & 1.71 & 1.68 & \textbf{1.72}  \\
			\hline
			FBANK (40-dim.)     & 1.08 & 1.08  & 1.14 & 1.09 & 1.10 & \textbf{1.10}  \\
			\hline
			FMLLR               & 0.67 & 0.68  & 0.71 & 0.66 & 0.67 & \textbf{0.68}  \\
			\hline
		\end{tabular}
	}
	\label{synops}
\end{table}

\pagebreak
\section*{Figure}

\end{document}